\documentclass[lettersize,journal]{IEEEtran}
\usepackage{amsmath,amsfonts}
\usepackage{algorithmic}
\usepackage{array}
\usepackage[caption=false,font=normalsize,labelfont=sf,textfont=sf]{subfig}
\usepackage{textcomp}
\usepackage{stfloats}
\usepackage{url}
\usepackage{verbatim}
\usepackage{graphicx}
\usepackage{cite}
\usepackage[pagebackref=true,breaklinks=true,letterpaper=true,colorlinks,bookmarks=false]{hyperref}

\usepackage{multirow}
\usepackage{xcolor}
\usepackage{amsthm,amsmath,amssymb}
\usepackage{mathrsfs}
\usepackage{gensymb}
\usepackage{makecell}
\usepackage[linesnumbered,ruled,vlined]{algorithm2e}

\begin{document}

\title{Cascade Forgery Mining Network for Fingerprint Presentation Attack Detection}

\author{Hongyan Fei, Chuanwei Huang, Zheng Wang, Pengcheng Luo, Jingwei Li and Jufu Feng*
\thanks{The correspondence author is Jufu Feng and the E-mail: fengjf@pku.edu.cn}
\thanks{Hongyan Fei, Chuanwei Huang, Zheng Wang, Pengcheng Luo, Jingwei li and Jufu Feng are with the State Key Laboratory of General Artificial Intelligence, School of Intelligence Science and Technology, Peking University, Beijing 100871, China.
E-mail: \{hongyanfei, wangzh, luopengcheng, lijingwei\}@stu.pku.edu.cn, \{huangcw, fengjf\}@pku.edu.cn\}.
}
}

\markboth{Journal of \LaTeX\ Class Files,~Vol.~14, No.~8, August~2021}%
{Shell \MakeLowercase{\textit{et al.}}: Fingerprint Presentation Attack Detection by Region Decomposition}


\maketitle
\begin{abstract}
Fingerprint Presentation Attack Detection (PAD) is a critical component of fingerprint identification systems, serving as a protective measure against unauthorized access. In this paper, we observe that different regions of a fingerprint image can exhibit varying Artifact Extraction Difficulty (AED), with high-AED regions requiring more sophisticated extraction mechanisms to capture more subtle discriminative evidence. To address this issue, we propose to quantify AED using local Gabor feature certainty and partition fingerprint images into multiple regions based on their respective AED values. We then propose an AED guided Cascade Forgery Mining Network (CFM-Net) that employs an adaptive-depth feature extraction architecture to detect more precise and comprehensive artifact evidence across regions with heterogeneous AED values. Furthermore, we introduce an Orientation Guided Adversarial Training (OGAT) module to filter out identity information from PAD features while preserving the integrity of original artifact evidence. Experimental evaluations on LivDet datasets demonstrate the superior performance of our approach compared to state-of-the-art methods and achieve significant improvement in the classification ability of high AED fingerprints.
\end{abstract}


\section{Introduction}
Fingerprint authentication has gained widespread adoption in security systems and applications owing to its convenience and dependable nature.  Nevertheless, the emergence of synthetic materials employed to replicate fingerprints, often termed as ``spoof'', has become a prevalent method for carrying out Presentation Attacks (PAs).  These spoof methods present a notable risk to automatic fingerprint authentication systems.

Fingerprint spoof methods can be categorized into consensual and non-consensual approaches based on the means of obtaining the source fingerprint. Traditional consensual methods like gummy~\cite{matsumoto2002impact}, using materials such as gelatin, eco flex, wood glue, and liquid latex to craft 2D or 3D replicas~\cite{arora2016design,arora2017gold}. Altered fingerprints \cite{yoon2012altered} and cadaver fingers~\cite{marasco2014survey} are also commonly utilized in spoof methods. On the other hand, non-consensual approaches like screenspoof~\cite{casula2021livdet} involve capturing latent fingerprints left on smartphone screens without user awareness. Spoof fingerprints made from this kind of approach are much more complex, but still pose an apocalyptic threat to fingerprint recognition systems.

Fingerprint authentication systems incorporate hardware and software techniques for Presentation Attack Detection (PAD) to address diverse PAs. However, due to the high cost of specific sensors utilized in hardware-based methods, software-based methods have become a more prevalent choice. 
Previous studies utilized conventional approaches to extract handcrafted features like curvelet and fractional Fourier transforms~\cite{lee2009fake,nikam2008fingerprint,xia2018novel}. With the development of deep learning, CNN-based methods~\cite{huang2023fingerprint,yook2024attention,reza2025cross} achieve state-of-the-art performance. Liu~\textit{et al.}~\cite{liu2022fingerprint} identified effective channels and noise channels within the network, suppressing the propagation of noise channels in feature maps to enhance generalization capability. Fei~\textit{et al.}~\cite{fei2024fingerprint} proposed that artifact evidence of fingerprints results from various causes, and decomposed fingerprint into ridge area and edge area for separate processing.

\begin{figure}[tbp]
\centering
\setlength{\abovecaptionskip}{0.05cm}
   \includegraphics[width=1\linewidth]{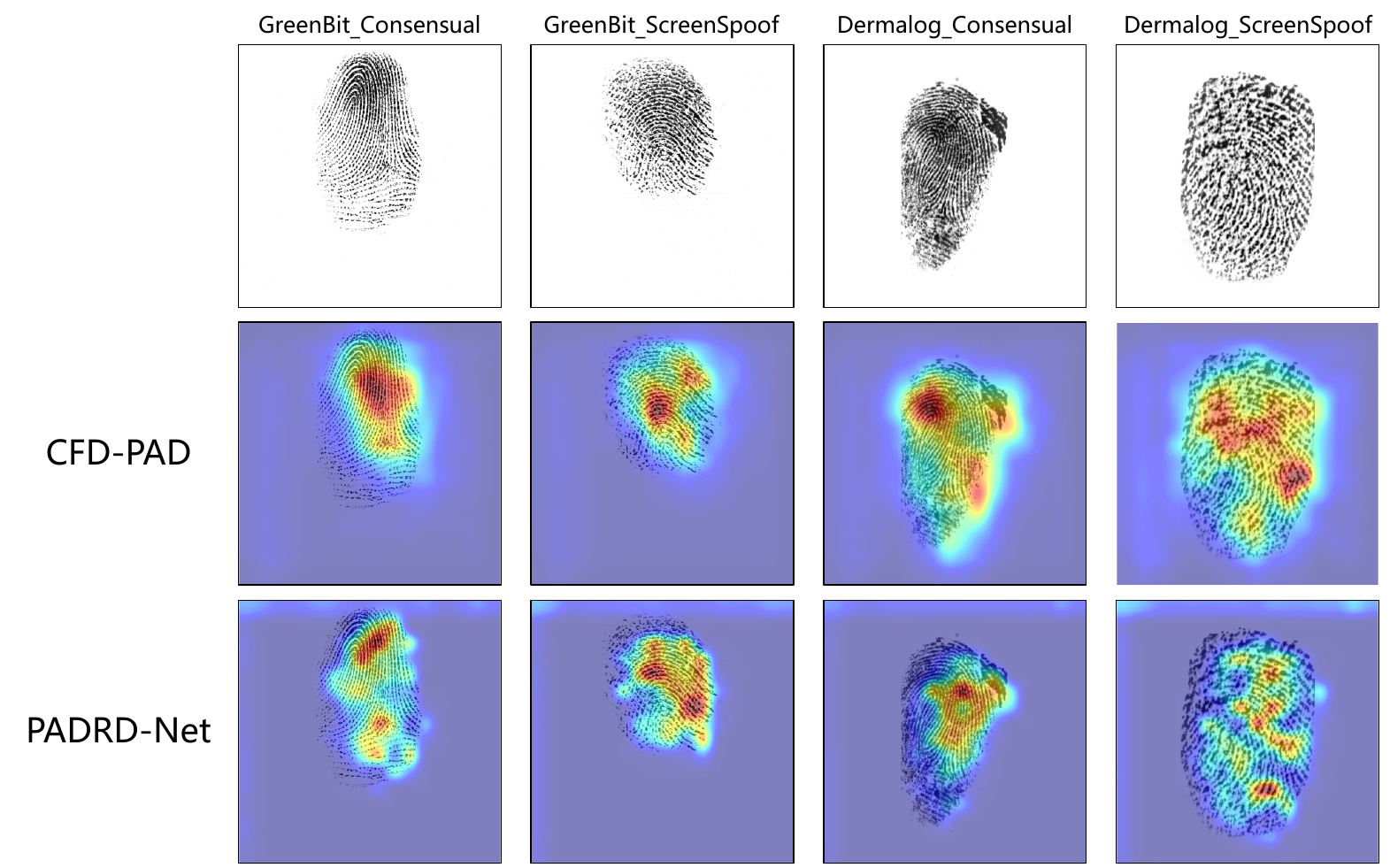}
   \caption{The CAM visualization of the current PAD methods PADRD-Net~\cite{fei2024fingerprint} and CFD-PAD~\cite{liu2022fingerprint} on LivDet2021 dataset.}
\label{fig:first}
\end{figure}

However, Class Activation Mapping (CAM)~\cite{zhou2016learning} visualization of some existing PAD methods (as shown in Fig.~\ref{fig:first}) reveals that they predominantly rely on partial fingerprint regions for decision-making. To investigate whether the neglected regions contain discriminative information, we conduct a comparative experiment using two parameter configurations: (1) the original PADRD-Net trained on original fingerprint images, and (2) the same network retrained exclusively on the less-attentional regions (i.e., regions with CAM $\le$ 0.2) cropped from the same training set. Both models are then evaluated on the less-attentional regions extracted from the test set under configuration (1). As shown in Table~\ref{tab:BOCM_validation}, these less-attentional regions exhibit minimal discriminability under the original parameters but achieve significant discriminative capability with the retrained parameters. These results demonstrate that discriminative artifact evidence exists in previously neglected regions but remains overlooked, thereby limiting the overall performance. 

This phenomenon can be attributed to the inherent variations in local ridge structure clarity across fingerprint regions. Recent analysis indicates that artifact evidence primarily resides within ridge details~\cite{fei2024fingerprint}. Consequently, regions with less distinct ridge structure naturally exhibit more subtle artifact patterns, while regions with prominent ridge structure display more pronounced traces. We term this spatial variation in extraction complexity as Artifact Extraction Difficulty (AED). Such AED-induced heterogeneity poses a significant challenge for rigid feature extraction architectures, which lack the adaptability to capture artifact evidence across regions with varying AED values.

To address this issue, we propose first assessing AED across fingerprint regions, then focusing on regions with higher AED for targeted processing. Based on the above analysis, a method for assessing AED should capture ridge structure clarity while remaining insensitive to intra-ridge details, as such fine-grained textures represent intrinsic properties of the evidence itself rather than variations in AED. Specifically, we propose that AED can be quantified as the sum of the top 9 response probabilities from a group of 90 discrete Gabor filters, where the orientation parameter is uniformly sampled from $0^\circ$ to $180^\circ$, and the frequency is set to match the average ridge period of fingerprint (e.g. 1/8 at 500 dpi). This fixed-frequency design ensures insensitivity to fine intra-ridge textures, while the aggregation of dominant orientations effectively quantifies local ridge structure clarity.

After obtaining AED, we propose an AED guided Cascade Forgery Mining Network (CFM-Net). The core of CFM-Net is an adaptive-depth architecture designed to mine artifact evidence across regions with varying AED. In practice, it processes input features through a cascade of stages, where each stage applies a dynamically generated binarized AED mask (with progressively increasing thresholds) to selectively route high-AED regions into deeper and more complex feature extraction stages to capture more subtle evidence. Features from all stages are then aggregated for the final classification, which allows for the integration of more comprehensive artifact evidence.

Furthermore, PAD methods often struggle with intrinsic coupling between fingerprint identity information and PAD features. Previous approaches such as MPRR operation~\cite{fei2024fingerprint} have been employed to destroy fingerprint identity information of the original input fingerprint image, resulting in the loss of original artifact evidence. To address this limitation, we propose an Orientation Guided Adversarial Training (OGAT) module that disentangles orientation information (i.e., identity-related features) from PAD features at the feature map level, thereby preserving the integrity of original artifact evidence.

The main contributions of this work are summarized as follows:

1) We observe that different regions of fingerprint images exhibit varying Artifact Extraction Difficulty (AED), and propose to quantify AED through local Gabor feature certainty for spatially-adaptive processing.

2) We propose an AED guided Cascade Forgery Mining Network (CFM-Net) that employs an adaptive-depth feature extraction architecture to capture more precise and comprehensive artifact evidence across regions with varying AED.

3) We introduce an Orientation Guided Adversarial Training (OGAT) module that filter out identity information from PAD features without compromising the integrity of artifact evidence.

4) Extensive experiments demonstrate that our proposed CFM-Net surpasses state-of-the-art PAD methods, achieving significant performance improvements particularly on fingerprints with high AED.

\section{Related Work}
\noindent\textbf{Hardware and Software-based Fingerprint PAD.} 
Fingerprint presentation attack detection has been approached using both hardware-based and software-based solutions. Hardware-based methods typically detect physiological liveness indicators such as blood flow~\cite{lapsley1998anti}, human odor~\cite{baldisserra2005fake}, and multi-perspective imaging~\cite{engelsma2018raspireader}, often utilizing specialized sensors including optical coherence tomography (OCT)-based systems~\cite{cheng2006artificial,zhang2024uniform} or multi-spectral sensors. Recently, Sun et al.~\cite{sun2023new} proposed the use of three hand-crafted OCT features to capture a more precise and detailed representation of the fingerprint’s internal physiological structure for enhanced PAD performance.

Software-based approaches perform PAD task without relying on specialized sensors. Instead, they utilize features extracted from captured fingerprint images. In earlier software techniques, manually designed features such as skin distortion information~\cite{antonelli2006fake}, fractional Fourier transforms~\cite{lee2009fake}, and curvelet transform-based methods~\cite{nikam2008fingerprint} were employed. However, recent advancements in deep learning have led to improved performance. For example, SlimResCNN~\cite{zhang2019slim} introduced a lightweight CNN with reduced processing time. Chugh~\textit{et al.}\cite{chugh2018fingerprint} extracted aligned localized minutia patches to highlight distinctive patterns. Liu~\textit{et al.}\cite{liu2021fingerprint} proposed a module that connects global and local modules, and the final score was computed by averaging the global and local spoof scores.
Furthermore, some contemporary methods~\cite{Gajawada2019UniversalMT,lekshmy2022one,rai2024open} employ transfer learning to generate synthetic patches on new sensors, materials, or both, thereby augmenting the available data.

\noindent\textbf{Adversarial representation learning.} 
Adversarial Representation Learning (ARL) is a machine learning technique applicable to both domain adaptation and domain generalization. It has been integrated into deep neural network architectures to extract distinctive representations for specific target prediction tasks while simultaneously filtering out some unwanted attributes inherent in the data~\cite{goodfellow2014generative,ganin2016domain,tzeng2017adversarial,zhang2018mitigating}. Grosz~\textit{et al.}~\cite{grosz2020fingerprint} leverage sensor label to learn sensor invariant representations for PAD task. In our approach, we employ the orientation derived from the fingerprint recognition system with adversarial training to remove identity information from the PAD information on the feature map level without compromising original artifact evidence.

\section{Methods}

\begin{figure*}[htbp]
\centering
\setlength{\abovecaptionskip}{0.05cm}
   \includegraphics[width=0.9\linewidth]{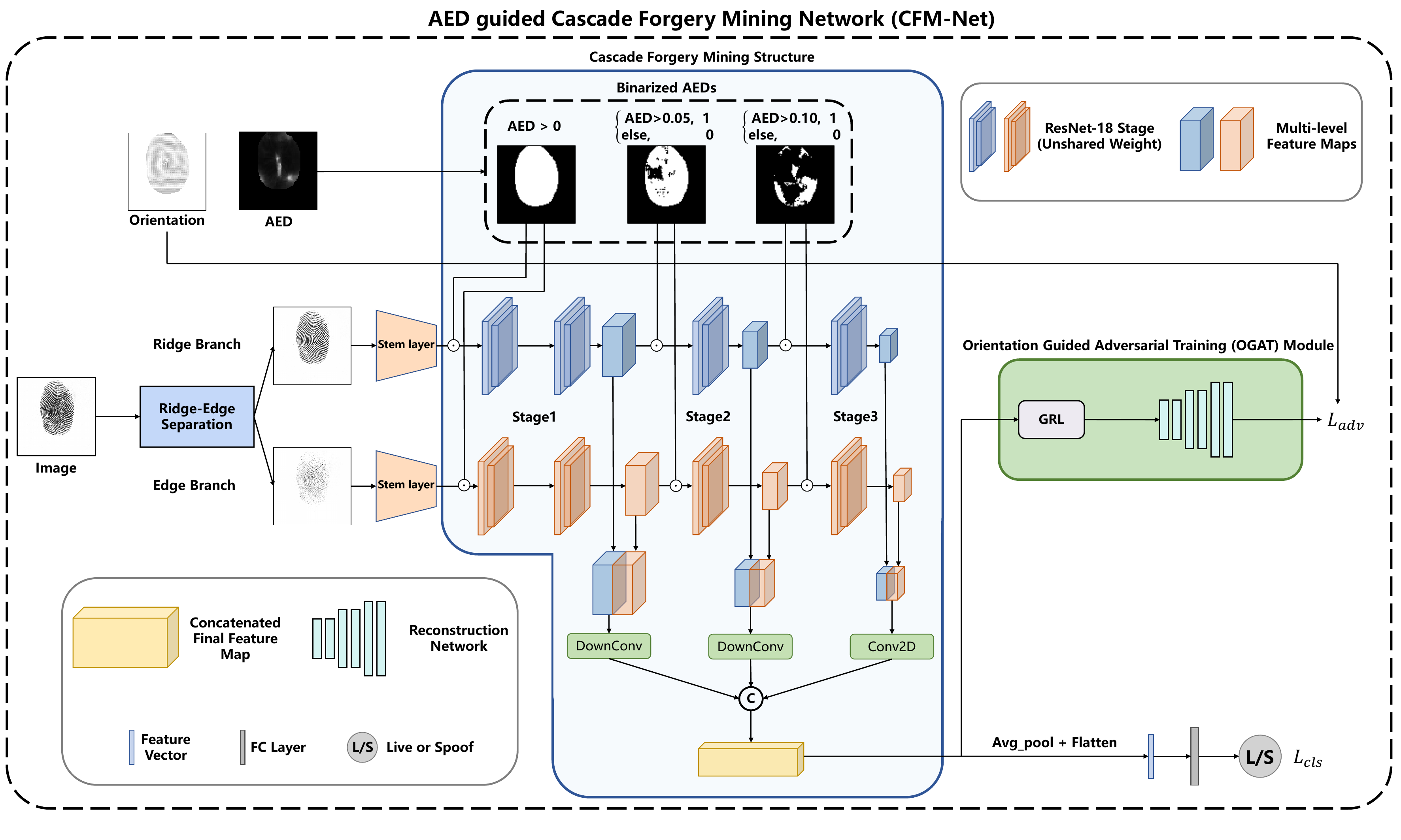}
   \caption{The overview of our proposed AED guided CFM-Net. Firstly, we split the input fingerprint into the ridge and edge areas. Then the ridge feature and edge feature are input into a dual-stream CFM structure, where the output features of each stage are concatenated to form the final feature map. Subsequently, the OGAT module is employed to filter out the identity information of the fingerprint before feeding the final feature map into the classifier.}
\label{fig:network}
\end{figure*}

In this paper, we observe that observe that different regions of fingerprint images exhibit varying AED. Subsequently, we develop CFM-Net that employs an adaptive-depth feature extraction architecture to capture more precise and comprehensive artifact evidence across regions with varying AED. Furthermore, we propose OGAT module that disentangles identity-related information from PAD features without compromising the integrity of the original artifact evidence. The overall pipeline of our proposed framework is illustrated in Fig.~\ref{fig:network}.

\subsection{preliminary}
Since the proposed AED quantification relies on local Gabor feature certainty, this section provides the necessary background on Gabor feature representation.

Due to the Gabor filters are widely employed in fingerprint image processing due to their optimal joint localization in both spatial and frequency domains~\cite{gabor1946theory}, making them particularly effective for enhancing and extracting the ridge structure of fingerprints~\cite{jain1997line,hong1998fingerprint}. These filters can selectively reinforce the periodic ridge-valley patterns while suppressing noise and non-relevant textures. The general two-dimensional Gabor filter kernel is defined as~\cite{hamamoto1998gabor}:
\begin{equation}
\begin{aligned}
h(x, y; f, \theta_{k}, \sigma_x, \sigma_y) = 
& \frac{1}{2\pi\sigma_x\sigma_y} 
\exp\left[-\frac{1}{2}\left(\frac{x_{\theta_k}^2}{\sigma_x^2} + \frac{y_{\theta_k}^2}{\sigma_y^2}\right)\right] \\
& \cdot \exp\left(i \cdot 2\pi f x_{\theta_k}\right), \qquad k = 1, \dots, m.
\end{aligned}
\end{equation}
where $x_{\theta_k} = x\cos\theta_k + y\sin\theta_k$ and $y_{\theta_k} = -x\sin\theta_k + y\cos\theta_k$. $f$ is the Central frequency, $m$ denotes the number of orientations, $\theta_k$ is the $k$-th orientation of the Gabor filter, and $\sigma_x$ and $\sigma_y$ are the standard deviations of the Gaussian envelope along the x and y axes. Since fingerprint ridge structures typically possess well-defined local frequency and orientation, the central frequency $f$ can be set to the reciprocal of the average inter-ridge distance, and the orientation parameter is commonly discretized as $\theta_k=\pi(k-1)/m, k=1,...,m$. The magnitude Gabor feature at the sampling point $(X, Y)$ can be defined as follows:
\begin{equation}
\label{gabor_2}
\begin{aligned}
G( X, Y, f, \theta_k, \sigma_x, \sigma_y ) 
= {} & \sum_{x = -w/2}^{w/2-1} \sum_{y = -w/2}^{w/2-1} I( X+x, Y+y ) \\
& \cdot h( x, y; f, \theta_k, \sigma_x, \sigma_y ),
\end{aligned}
\end{equation}
where \(I(\cdot,\cdot)\) denotes the pixel gray-level value, and \(w\) is the window size. Once the Gabor filter parameters are fixed, \(m\) different \(w \times w\) Gabor filter kernels are constructed. Fingerprint image is then convolved with these \(m\) kernels, yielding \(m\) corresponding Gabor responses. Consequently, a \(w \times w\) window is effectively compressed into \(m\) discriminative Gabor features.

\subsection{Artifact Extraction Difficulty Analysis}

As illustrated in Fig.~\ref{fig:first}, CAM visualization reveals that current methods predominantly concentrate on partial regions of the fingerprint, raising questions about the informative value of neglected regions. To investigate whether these overlooked regions contain discriminative artifact evidence, we conduct a comparative experiment on LivDet 2021 dataset. These regions are evaluated under two distinct parameter configurations of PADRD-Net structure: (1) the original PADRD-Net trained on original fingerprint images, and (2) the same network retrained exclusively on the less-attentional regions (i.e., regions with CAM $\le$ 0.2) cropped from the same training set. Both models are then evaluated on the less-attentional regions extracted from the test set under configuration (1).

Table~\ref{tab:BOCM_validation} demonstrates that less-attentional regions exhibit limited discriminability under the original parameters (Configuration 1), yet achieve effective classification with region-specific parameters (Configuration 2). This contrast reveals that discriminative artifact evidence indeed exists in less-attentional regions, thereby limiting their overall performance.

Recent analysis of artifact evidence~\cite{fei2024fingerprint} indicates that artifact evidence primarily resides within ridge details, such as the secretion of sweat during contact with the capture sensor, or abnormal burrs resulting from the properties of spoof materials and imperfections in manufacturing precision. Consequently, regions with less distinct ridge structure naturally exhibit more subtle artifact evidence patterns, while areas with clearer ridge structure display more pronounced patterns. We term this spatial variation in extraction complexity as Artifact Extraction Difficulty (AED). The resulting heterogeneity poses a significant challenge for rigid feature extraction architectures, which struggle to adaptively capture artifact evidence across regions with varying AED values.

\begin{table*}[htbp]
\caption{Performance comparison using ``original'' vs. ``region-specific'' parameter configurations. ``original'' means the original PADRD-Net trained on original fingerprint images, and ``region-specific'' means the same network retrained exclusively on the less-attentional regions cropped from the same training set.}
\setlength{\tabcolsep}{1pt}
\begin{center}
    \begin{tabular}{cccccc}
    \hline
    \multirow{2}{*}{Parameters} & \multirow{2}{*}{test data} & \multicolumn{4}{c}{LivDet 2021} \\
\cline{3-6}          &       &ACC(\%)$\uparrow$ &ACER(\%)$\downarrow$   &BPCER@APCER=$1\%$(\%)$\downarrow$    &TDR@FDR=$1\%$(\%)$\uparrow$    \\
    \hline
    original parameters & \multirow{2}{*}{less-attentional regions} &57.39 &45.62 & 91.02      & 4.39 \\
    region-specific parameters &       &85.17 &14.32 & 44.36      & 54.68\\
    \hline
    \end{tabular}%
\end{center}
    \label{tab:BOCM_validation}
\end{table*}%

To address this issue, we propose a framework that first quantifies the AED across different fingerprint regions, then subsequently focuses on regions with higher AED for targeted processing. The preceding analysis posits that artifact evidence is more easily extracted in regions with clearer ridge structure. Consequently, a method for assessing AED must indicate ridge structure clarity while remaining insensitive to intra-ridge details, as such details primarily represent inherent characteristics of the evidence itself rather than variations of AED. Given that Gabor features can represent ridge structure, we propose using local Gabor feature certainty to quantify AED. 
Specifically, based on Eq.~\ref{gabor_2}, the frequency parameter $f$ is set to match the average ridge period of fingerprint (e.g. 1/8 at 500 dpi) and the orientation parameter $\theta$ to range from $0^\circ$ to $180^\circ$ with a step size of $2^\circ$, yielding a Gabor filter group consisting of 90 filters. We set the window size $w$ to three times of inter-ridge distance and convolve the Gabor filter group with the input fingerprint image of size $H \times W \times 1$ using a stride of 8, obtaining a Gabor feature $\left\{ {{G(x, y, \theta_i)}} \right\}_{i = 0}^{89}$ with dimensions $H/8 \times W/8 \times 90$. Each 90-dimensional vector at a spatial position represents the local Gabor feature of the corresponding image region. We quantify local Gabor feature certainty by computing the sum of the top 9 highest probabilities. Since lower Gabor feature certainty indicates less distinct ridge structure and thus corresponds to higher AED, the AED at position $(X, Y)$ can be expressed as:
\begin{equation}
\label{orientation_mask}
\begin{aligned}
{AED}(X, Y) &= 1 - \sum_{j=0}^{8}G'(x, y, \theta_j), \\
\left\{ {{G'(x, y, \theta_j)}} \right\}_{j = 0}^{89} &= \mathrm{Sorted}\left(\mathrm{softmax}\left( \{ \left\{ {{G(x, y, \theta_i)}} \right\}_{i = 0}^{89} \right)\right)_{\downarrow}.
\end{aligned}
\end{equation}

\begin{figure}[htbp]
\centering
\setlength{\abovecaptionskip}{0.05cm}
   \includegraphics[width=1\linewidth]{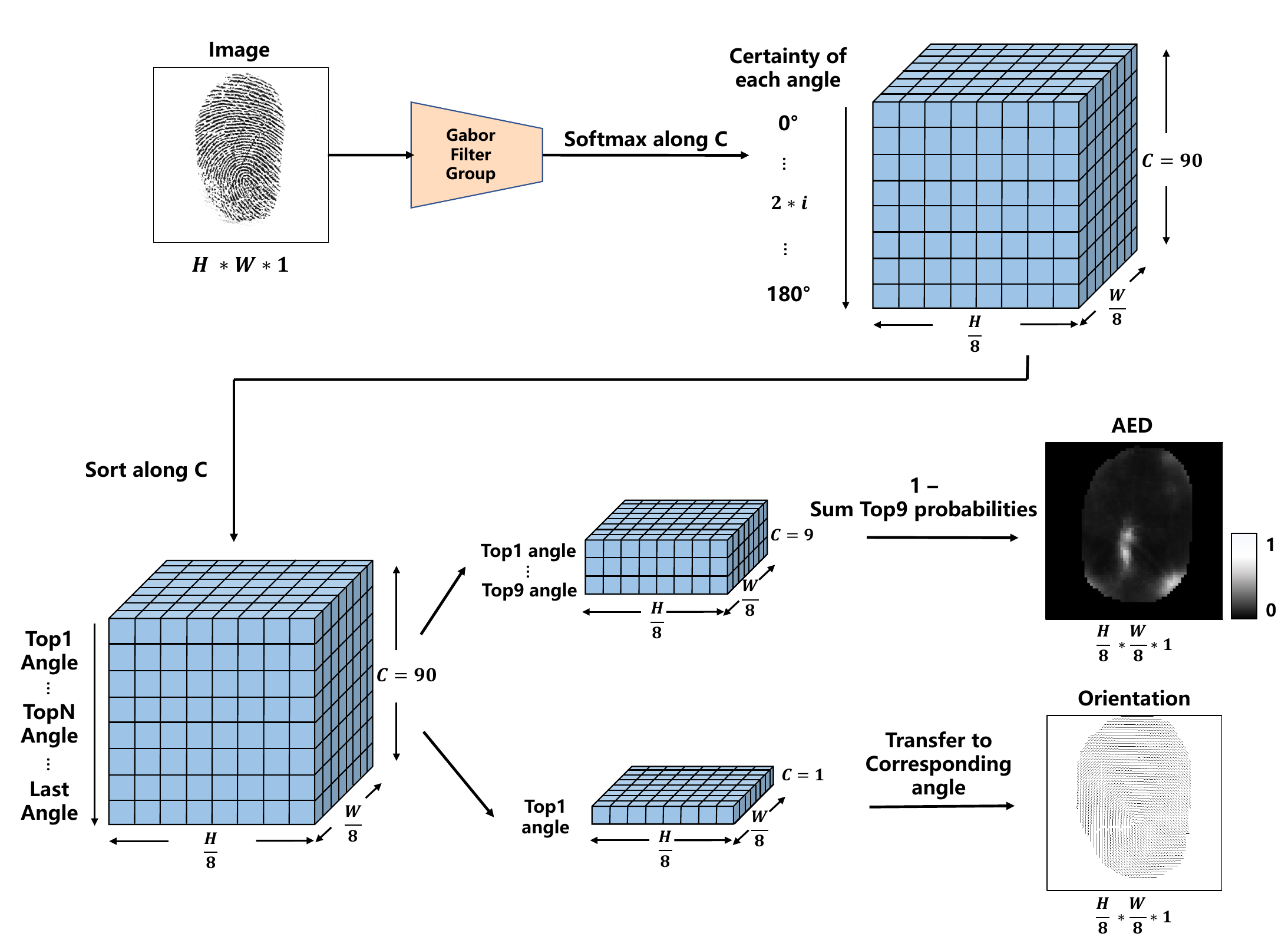}
   \caption{The generation process of AED is generated based on the sum of top 9 probabilities of 90 discrete angles of fingerprint local Gabor feature certainty.}
\label{fig:OCM}
\end{figure}

The computational process of AED is illustrated in Fig.~\ref{fig:OCM}. Overall, quantifying AED using the sum of the top 9 highest probabilities of local Gabor features offers three key advantages:

\noindent1. The fixed-frequency Gabor filter group responds to variations in ridge structure clarity while maintaining insensitivity to intra-ridge details, thereby ensuring that the extracted features align with the physical requirements of AED.

\noindent2. By utilizing the sum of the top 9 probabilities rather than only the highest probability, the method enhances robustness in regions with sharp orientation changes, such as those containing minutiae. In these regions, the certainty of the top 1 probability may decrease even though the ridge structure remains clear.

\noindent3. The Gabor filter group enables simultaneous estimation of both AED and the orientation field. Since orientation field estimation is typically required in fingerprint authentication systems, this approach reduces the computational overhead of integrated Authentication-PAD systems in practical scenarios.

\begin{figure*}[htbp]
\centering
\setlength{\abovecaptionskip}{0.05cm}
   \includegraphics[width=0.7\linewidth]{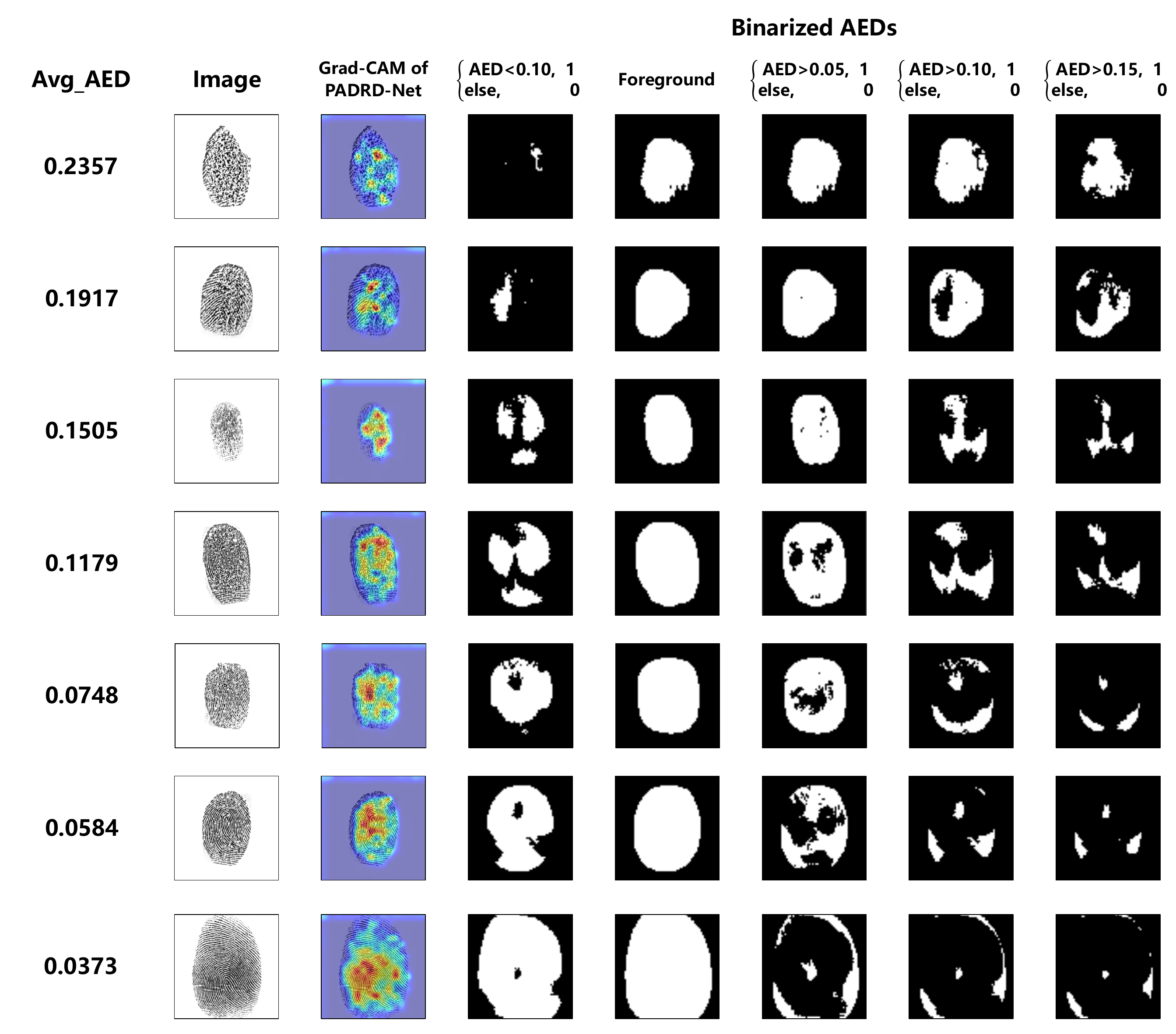}
   \caption{Visualization of fingerprints with different average values of AED, CAM results for current state-of-the-art method PADRD-Net~\cite{fei2024fingerprint}, and binarized AEDs with different thresholds. It is evident that the decision regions resemble those with AED values below 0.10.}
\label{fig:mask}
\end{figure*}

\begin{table*}[htbp]
  \caption{The performance of current PAD methods on fingerprints with different ranges of average AED of LivDet 2021 dataset.}
\setlength{\tabcolsep}{6pt}
\begin{center}
    \begin{tabular}{cccc}
    \hline
    \multirow{2}{*}{Methods} &\multirow{2}{*}{Test Data} & \multicolumn{2}{c}{LivDet2021} \\
\cline{3-4} & &BPCER@APCER=$1\%$(\%)$\downarrow$    &TDR@FDR=$1\%$(\%)$\uparrow$    \\
    \hline
    \multirow{3}{*}{FSB\cite{chugh2018fingerprint}} & Avg\_AED $\leq$ 0.10 &  33.38     & 59.92 \\
     & 0.10 $\textless$ Avg\_AED $\textless$ 0.15 & 43.64 & 58.66\\
     & Avg\_AED $\geq$ 0.15 &74.64 &36.56 \\
     \hline
    \multirow{3}{*}{CFD-PAD~\cite{liu2022fingerprint}} & Avg\_AED $\leq$ 0.10 & 72.87    & 54.79\\
     & 0.10 $\textless$ Avg\_AED $\textless$ 0.15 & 91.90 & 44.25\\
     & Avg\_AED $\geq$ 0.15 & 95.73 & 38.26\\
     \hline
    \multirow{3}{*}{PADRD-Net~\cite{fei2024fingerprint}} & Avg\_AED $\leq$ 0.10 & 23.25      & 83.92 \\
     & 0.10 $\textless$ Avg\_AED $\textless$ 0.15 & 28.18 & 83.36 \\
     & Avg\_AED $\geq$ 0.15 & 77.46 & 27.96\\
    \hline
    \end{tabular}%
\end{center}
    \label{tab:Low-qua_1}
\end{table*}%

We conduct a visualization experiment as illustrated in Fig.~\ref{fig:mask}. The first column presents the average AED computed across the foreground region of each fingerprint, while the third column displays CAM visualization results obtained from the state-of-the-art PADRD-Net. The remaining columns show binarized AED maps at different threshold levels. A clear correspondence is observed between the attention regions highlighted by CAM and regions where the AED value is below 0.10, suggesting that this method primarily captures artifact evidence in low-AED regions.

To further validate whether this tendency generalizes to existing methods, we design a comparative experiment presented in Table~\ref{tab:Low-qua_1}. We separate original fingerprint images from the LivDet2021 test set into multiple subsets based on varying ranges of average AED values and evaluate performance using BPCER@APCER=$1\%$ and TDR@FDR=$1\%$ metrics. The results demonstrate that current methods consistently exhibit substantial performance degradation when processing fingerprints with higher average AED. Consequently, leveraging AED information to guide the extraction of more comprehensive artifact evidence represents a promising direction worthy of further investigation.

\subsection{CFM-Net}

Following the quantification of AED, we partition each fingerprint image into multiple regions according to their respective AED values, enabling focused processing of high-AED regions. Specifically, we propose an AED guided Cascade Forgery Mining Network (CFM-Net) that employs an adaptive-depth architecture to mine more comprehensive artifact evidence.

The structure of CFM-Net is illustrated in Fig.~\ref{fig:network}, CFM-Net first splits the input fingerprint into the ridge and edge areas. This separation is motivated by the fact that artifacts in these areas originate from different mechanisms (contact-based and non-contact-based)~\cite{fei2024fingerprint}. Each area is then processed by a dedicated branch within a two-branch backbone network. The core of each branch is our Cascade Forgery Mining (CFM) structure. Prior to each stage in a branch, a binarized AED map is generated using a progressively increasing threshold, which is used to mask the input features. This adaptive masking mechanism dynamically routes features: regions with higher AED values are progressively emphasized and processed by deeper and more complex modules within the cascade structure, while lower-AED regions undergo lighter processing. Thus, the network achieves AED-focused depth adaptation. Finally, hierarchical feature representations from all stages and both branches are aggregated and passed to a classifier for the final prediction. 

Algorithm.~\ref{CM_alg} delineates the workflow of the CFM structure within a single branch. Compared to employing multiple independent adaptive-depth feature extractors for regions with varying AED levels, the CFM structure utilizes shared shallow stages to maintain foundational feature interaction across the entire fingerprint region. This design prevents the loss of artifact evidence located at regional boundaries while simultaneously reducing computational cost through parameter sharing in the shallow layers.

\begin{algorithm}[htbp]
\SetAlgoLined
\scriptsize
\textbf{Input:} Data $X$, binarized AED masks $\{ M_1, M_2, \ldots, M_N \}$ \\
\tcp{M progressively emphasizes regions with higher AED.}
\textbf{Network:} Multiple stages $\{ S_{stem}, S_1, S_2, \ldots, S_N, S_{cls} \}$ \\
\textbf{Output:} Live/Spoof prediction $Y$ \\
Initialization: $F_0 = S_{stem}(X)$; $F_{concat} \leftarrow \emptyset$.\\
\For{$n = 1$ to $N$}{
    Downsample $M_n$ to $F_{n-1}$ size \\
    $F_n = S_n(F_{n-1} \odot M_n)$ 
}
\For{$n = 1$ to $N$}{
    DownConv $F_n$ to $F_N$ size \\
    $F_{concat} \leftarrow F_{concat} \Vert F_n$ \tcp{Concatenate along channel dimension}
}
$Y = S_{cls}(F_{concat})$ 
\caption{AED guided Cascade Forgery Mining (CFM) Structure}
\label{CM_alg}
\end{algorithm}

\subsection{Orientation Guided Adversarial Training Module}

Traditional fingerprint PAD methods face challenges in decoupling PAD feature and identity information, due to the fact that both are intrinsically embedded within ridge-valley structure. Fei~\textit{et al.}~\cite{fei2024fingerprint} proposed the MPRR operation to suppress fingerprint identity information through random rotation of minutiae patches. However, a fundamental limitation of such image-level approaches is their reliance on explicit perturbation of the ridge structure, which inevitably compromises the integrity of original artifact evidence.
To address this limitation, we propose an Orientation Guided Adversarial Training (OGAT) module that filter out identity information at the feature map level rather than at the image level. Specifically, leveraging the fact that orientation field is the identity-related feature, we design an orientation reconstruction network comprising 6 convolutional layers trained with adversarial loss to process features extracted by the main PAD network. A Gradient Reversal Layer (GRL)~\cite{ganin2015unsupervised} is then employed to enforce inverse optimization dynamics between the main PAD network and the reconstruction network. This adversarial training process can be formulated as:
\begin{align}
\label{orientation mask}
\nonumber
&\min_{\theta_{ogat}} \max_{\theta_{main}} \mathcal{L}_{adv}\\
&=\| M_{ogat}(M_{main}(I,\theta_{main}),\theta_{ogat}) - O_{gt} \|_{1}
\end{align}
where $I$ denotes the input fingerprint image, $M_{main}$ and $\theta_{main}$ represent the model and parameters of the main PAD network, $M_{ogat}$ and $\theta_{ogat}$ denote the model and parameters of the OGAT module, and $O{gt}$ represents the fingerprint orientation field. The adversarial loss is computed as the L1 distance between the OGAT output and $O_{gt}$. As illustrated in Fig.~\ref{fig:network}, the OGAT module eliminates identity information from learned representations while preserving the discriminative artifact evidence essential for PAD task without harming the original artifact evidence.

\subsection{Loss Function}
Since the difference between spoofs might be much larger than live ones, we adopt the PA-Adaptation loss~\cite{liu2022fingerprint} to alter the feature distribution by pulling all the live fingerprints close while pushing the spoof ones of different attack types apart. The loss function is given by:
\begin{equation}
\mathcal{L}_{padp} = \{\begin{array}{ll}
1 - \cos \left( {{a_i},{b_i}} \right), & \text{if } y _i = 1 \\
\max \left( {0,\cos \left( {{a_i},{b_i}} \right) - \alpha } \right), & \text{if } y _i = -1
\end{array}
\end{equation}
where $\left\{{a_i},{b_i}\right\}$ indicates the samples' feature vector in a mini-batch. When $a_i$ and $b_i$ belong to the same material (including live skin), they are considered positive pairs, with $y_i=1$. When $a_i$ and $b_i$ belong to different materials, they are considered negative pairs, with $y_i=-1$. $\alpha$ is an enforced margin of negative pairs' cosine value, we set $\alpha=0$ in our method.
The final loss function of the CFM-Net is the weighted summation of three losses:
\begin{equation}
    \mathcal{L}_{all} = \mathcal{L}_{cls} + \mathcal{L}_{padp} + \lambda \cdot \mathcal{L}_{adv},
    \label{eq:loss_all}
\end{equation}
where $\mathcal{L}_{cls}$ is the cross entropy loss for live/spoof binary classification, $\mathcal{L}_{padp}$ is the PA-Adaptation loss, $\mathcal{L}_{adv}$ is the adversarial training loss, and $\lambda$ is the coefficient of balance.

\section{Experiments}
In this section, we conduct extensive experiments on fingerprint PAD tasks to demonstrate the effectiveness of our method. We introduce the experimental settings (Sec.~\ref{section:Experimental Settings}), experimental comparisons on LivDet datasets (Sec.~\ref{section:Experimental Comparisons}---\ref{sec: Performance on Fingerprints with Different Qualities}), and a series of ablation studies and analysis experiments (Sec.~\ref{section:Effectiveness Validation of Each Proposed Module}---\ref{section:Future Work}) in sequence.

\subsection{Experimental Settings}
\label{section:Experimental Settings}

\noindent\textbf{Datasets.} We evaluate our method on the LivDet 2021~\cite{casula2021livdet} and LivDet 2019~\cite{orru2019livdet} datasets, which represent the benchmarks from the two LivDet competitions. Table~\ref{tab:dataset_21} and Table~\ref{tab:dataset_19} provide comprehensive overviews of these datasets. A critical characteristic of both benchmarks is that the spoof materials used in the training and test sets are completely disjoint, ensuring rigorous evaluation of generalization capability. LivDet 2021 introduces additional complexity by incorporating the non-consensual ScreenSpoof attack method in its test set. Throughout subsequent tables, "CC" denotes the consensual capture subset, while "SS" refers to the ScreenSpoof subset. LivDet 2019 distinguishes itself by being the first competition to include multi-material composite spoofs, which combine materials of varying physical consistencies and chemical compositions, thereby presenting a more challenging and realistic attack scenario.

\begin{table*}[htbp]
\caption{The overview of LivDet 2021 dataset.}
\setlength{\tabcolsep}{6pt}
\begin{center}
    \begin{tabular}{cccccccccc}
        \hline
        \multirow{2}{*}{Dataset}      & \multicolumn{3}{c}{Training} & \multicolumn{6}{c}{Test (Consensual/ScreenSpoof)} \\
        \cline{2-10}
         & \multicolumn{1}{l}{Live} & \multicolumn{1}{l}{Latex} & \multicolumn{1}{c}{PRroFast} & Live  & \multicolumn{1}{c}{Mix 1} & \multicolumn{1}{l}{BodyDouble} & \multicolumn{1}{c}{ElmersGlue} & GLS20 & RFast30 \\
        \hline
        GreenBit & 1250  & 750   & 750   & 2050/2050 & 820/820 & 820/820 & 820/820 &-       &-  \\
        Dermalog & 1250  & 750   & 750   & 2050/2050 &-       &-       &-       & 1230/1230 & 1230/1230 \\
        \hline
        \end{tabular}%
\end{center}
  \label{tab:dataset_21}%
\end{table*}%

\begin{table*}[htbp]
\caption{The overview of LivDet 2019 dataset.}
\setlength{\tabcolsep}{5pt}
\begin{center}
    \begin{tabular}{ccccccccccc}
    \hline
    \multirow{2}{*}{Dataset} & \multicolumn{6}{c}{Training}                 & \multicolumn{4}{c}{Test} \\
    \cline{2-11}
          & \multicolumn{1}{c}{Live} & \multicolumn{1}{c}{Wood Glue} & \multicolumn{1}{c}{Ecoflex} & \multicolumn{1}{c}{Body Double} & \multicolumn{1}{c}{Latex} & \multicolumn{1}{c}{Gelatine} & \multicolumn{1}{c}{Live} & \multicolumn{1}{c}{Mix 1} & \multicolumn{1}{c}{Mix 2} & \multicolumn{1}{c}{Liquid Ecoflex} \\
\hline    
GreenBit & 1000  & 400   & 400   & 400   &-      &-      & 1020  & 408   & 408   & 408 \\

    Orcanthus & 1000  & 400   & 400   & 400   &-      &-     & 990   & 384   & 308   & 396 \\

    DigitalPersona & 1000  & 250   & 250   &-       & 250   & 250   & 1019  & 408   & 408   & 408 \\
    \hline
    \end{tabular}%
\end{center}
  \label{tab:dataset_19}%
\end{table*}%

\noindent\textbf{Metrics.} Adhering to ISO/IEC 30107-3:2023 standards, we assess classification performance using the Bona fide Presentation Classification Error Rate at a fixed Attack Presentation Classification Error Rate (BPCER@APCER=1\%) and the True Detection Rate at a fixed False Detection Rate (TDR@FDR=1\%). The APCER represents the proportion of presentation attacks incorrectly classified as bona fide presentations within a given scenario, while the BPCER signifies the ratio of bona fide presentations incorrectly classified as presentation attacks within the same scenario. Specifically, in the context where live/spoof samples are considered as positive/negative, their computational formulas are as follows:
\begin{gather}
APCER = \frac{FP}{TN + FP},\\
BPCER = \frac{FN}{TP + FN}.
\end{gather}

It should be noted that the LivDet competition reviews did not originally include metrics such as BPCER@APCER and TDR@FDR. The metrics available were limited to Ferrlive and Ferrfake (corresponding to BPCER and APCER), along with overall ACC. Therefore, in our experiments, we introduce the Average Classification Error Rate (ACER) metric to facilitate comparisons between LivDet competition methods and our proposed approach. The ACER is formulated as follows:
\begin{equation}
\label{ACER cal}
    ACER = \frac{APCER + BPCER}{2}.
\end{equation}

We also employ the Accuracy (ACC) metric on the overall dataset to directly reflect the accuracy of model classification.

The lower values of ACER and BPCER@APCER=$1\%$ indicate better performance of the PAD model, while higher values of TDR@FDR=$1\%$ and ACC signify an improved performance of the PAD model.

\noindent\textbf{Implementation Details.} We set the coefficient of balance $\lambda$ to 1. During training, we employ the Adam optimizer and decrease the learning rate from 1e-4 to 5e-6 using the CosineAnnealingLR scheduler. 

\subsection{Experimental Comparisons}
\label{section:Experimental Comparisons}

In this section, we compare our proposed CFM-Net with state-of-the-art methods and LivDet competition winners across three evaluation scenarios. We select megvii\_ensemble~\cite{casula2021livdet} from the LivDet 2021 competition and PADUnkFv~\cite{orru2019livdet} from the LivDet 2019 competition, as they achieved the highest live/spoof classification accuracy on the test sets according to competition reviews. By providing a comprehensive comparison, we aim to demonstrate the superiority of our proposed method in terms of ACC, ACER, BPCER@APCER=$1\%$, and TDR@FDR=$1\%$.

\textbf{Cross-Material Evaluation.} Given that capture sensors are typically controllable in deployed systems, cross-material generalization represents the most practical evaluation scenario. We assess the cross-material robustness of CFM-Net following the official LivDet protocol, wherein spoof materials in the test set are completely disjoint from those in the training set. Quantitative results on LivDet 2021 and LivDet 2019 are presented in Table~\ref{tab:cross material} and Table~\ref{tab:cross material 19}, respectively. CFM-Net achieves superior performance on both benchmarks, surpassing all competing methods. Notably, on LivDet 2021, CFM-Net yields substantial improvements in BPCER@APCER=$1\%$ for the overall dataset, decreasing from 20.39\% to 14.60\% compared to previous state-of-the-art, while simultaneously attaining the best performance on each individual subset. We attribute these gains to two key architectural innovations: (1) the CFM structure enables adaptive-depth feature extraction tailored to regions with varying AED, facilitating more comprehensive capture of artifact evidence; (2) the OGAT module suppresses identity-related information while preserving discriminative original artifact evidence, thereby enhancing cross-material generalization capability.

\begin{table*}[htbp]
  \caption{Performance comparison among previous state-of-the-art methods and LivDet competition winner on LivDet 2021 dataset. It also can be regarded as Cross-Material scenario. BPCER refers to BPCER@APCER=1\%(\%). TDR refers to TDR@FDR=1\%(\%). The best results are highlighted.}
\begin{center}
\setlength{\tabcolsep}{2pt}
    \begin{tabular}{ccccccccccccccccc}
    \hline
    \multirow{3}{*}{Method} & \multicolumn{15}{c}{LivDet2021~~~~ACC(\%)$\uparrow$~~~~ACER(\%)$\downarrow$~~~~BPCER@APCER=1\%(\%)$\downarrow$~~~~TDR@FDR=1\%(\%)$\uparrow$~~~~} \\
\cline{2-17}          & \multicolumn{3}{c}{GreenBit\_CC} & \multicolumn{3}{c}{GreenBit\_SS} & \multicolumn{3}{c}{Dermalog\_CC} & \multicolumn{3}{c}{Dermalog\_SS} & \multicolumn{4}{c}{ALL} \\
\cline{2-17}          & ACER   & BPCER  &TDR   & ACER   & BPCER  &TDR   & ACER   & BPCER  &TDR   & ACER   & BPCER  &TDR  &ACC  & ACER   & BPCER & TDR\\
    \hline
    ResNet18  & 4.40        &8.40       & 62.18       &27.59       &35.79    &23.91       &7.28       & 11.74  &42.67  &38.81 &85.94 &12.48  &79.15  &19.51       & 63.77       &21.70\\

    FSB~\cite{chugh2018fingerprint}     &2.39       &6.29       &95.56       &15.89       &25.56       &63.98        &1.72       &4.58       &96.76       &42.92  &53.80  &13.41 &82.96  &15.73       &36.09  &68.01\\

    CFD-PAD~\cite{liu2022fingerprint}     &6.02       &9.21       &83.23       &26.14       &46.29       &23.84       &3.36       &4.97       &94.77       &51.19       &93.41  &7.12  &78.42   &21.67       &85.07  &30.42\\
    hallymMMC~\cite{casula2021livdet}     &19.89       &--       &--       &38.81       &--       &--       &4.78       &--       &--       &14.13  &-- &-- &86.32 &19.40 &-- &--\\
    megvii\_ensemble~\cite{casula2021livdet} &\textbf{1.39}       &--       &--       &6.84       &--       &--       &\textbf{0.26}       &--       &--      &14.45  &-- &-- &93.79 &11.40 &-- &--\\
    FPAD-SupCon~\cite{huang2023fingerprint} &3.65       &--       &--       &13.38       &--       &--       &1.66       &--       &--      &\textbf{2.62}  &-- &-- &95.02 &4.78 &16.22 &83.36\\
    PADRD-Net~\cite{fei2024fingerprint}     &4.70       &10.02    &87.12   &7.14      &19.56  &82.72   &2.19  &3.53      &97.26    &7.66   &37.78       &75.72   &94.30   &5.42       &20.39  &87.12\\

    CFM-Net (Ours)     &1.79       &\textbf{3.91}       &\textbf{97.30}         &\textbf{5.16}       &\textbf{16.96}   &\textbf{85.99}    &1.99       &\textbf{3.24}       &\textbf{97.96}       &8.98  &\textbf{30.73} &\textbf{78.70} &\textbf{95.34} &\textbf{4.48} &\textbf{14.60} &\textbf{88.91}\\
   \hline
    \end{tabular}%
\end{center}
    \label{tab:cross material}%
\end{table*}%

\begin{table*}[htbp]
  \caption{Performance comparison among previous state-of-the-art methods and LivDet competition winner on LivDet 2019 dataset. It also can be regarded as Cross-Material scenario. BER refers to BPCER@APCER=1\%(\%). TDR refers to TDR@FDR=1\%(\%). The best results are highlighted.}
\begin{center}
\setlength{\tabcolsep}{6pt}
\begin{tabular}{cccccccccccccc}
    \hline
    \multirow{3}{*}{Method} & \multicolumn{13}{c}{LivDet2019~~~~ACC(\%)$\uparrow$~~~~ACER(\%)$\downarrow$~~~~BPCER@APCER=1\%(\%)$\downarrow$~~~~TDR@FDR=1\%(\%)$\uparrow$~~~~} \\
\cline{2-14}          & \multicolumn{3}{c}{GreenBit} & \multicolumn{3}{c}{DigitalPersona} & \multicolumn{3}{c}{Orcanthus} & \multicolumn{4}{c}{ALL} \\
\cline{2-14}          & ACER   & BER  &TDR  & ACER   & BER  &TDR  & ACER   & BER  &TDR  &ACC & ACER   & BER  &TDR \\
    \hline
    ResNet18     &14.19       &27.38        &40.54       &28.02         &44.95       &26.76        & 9.26      &18.41 &75.63     &81.44  &17.42       &37.63        &43.43 \\

    FSB~\cite{chugh2018fingerprint}     &1.56       &1.76       & 97.79      &6.34       &10.40       &  52.04     &3.90       &9.09   &89.43  &96.03  &3.96   &8.98   &81.44   \\

    CFD-PAD~\cite{liu2022fingerprint}     & 14.44      &49.21       &33.02       &10.84       &35.23       &29.41       &10.35       &27.67   &87.05 &85.24 &11.88   &38.52  &33.90    \\

   PADUnkFv~\cite{orru2019livdet}     &2.40       &--       &--       &6.24       &--       &--       &2.83       &--    &-- &96.17 &3.82   &--  &--      \\

    PADRD-Net~\cite{fei2024fingerprint}     &1.40       &\textbf{0.98}   &99.26    &4.76       &9.42   &72.39   &3.03    &\textbf{1.11}  &98.07  &97.11 &3.07       &4.19       &92.48       \\
    CFM-Net (Ours)     &\textbf{1.32}       & 1.17  & \textbf{99.43}   & \textbf{3.72}      & \textbf{6.86}  & \textbf{92.44}  & \textbf{2.57}   & 2.02 & \textbf{98.37} & \textbf{97.57}   & \textbf{2.55}      &  \textbf{3.99}     & \textbf{95.54}      \\
   \hline
    \end{tabular}%
\end{center}
  \label{tab:cross material 19}%
\end{table*}%

\begin{table*}[htbp]
  \caption{Performance comparison with previous methods under Cross-Sensor scenario on LivDet 2021 dataset. BER refers to BPCER@APCER=1\%(\%). TDR refers to TDR@FDR=1\%(\%). The best results are highlighted.}
\setlength{\tabcolsep}{6pt}
\begin{center}
    \begin{tabular}{cccccccccccccc}
    \hline
    \multicolumn{2}{c}{\multirow{2}{*}{LivDet 2021}} & \multicolumn{12}{c}{LivDet2021~~~~ACER(\%)$\downarrow$~~~~BPCER@APCER=1\%(\%)$\downarrow$~~~~TDR@FDR=1\%(\%)$\uparrow$~~~~} \\
\cline{3-14}    \multicolumn{2}{c}{} & \multicolumn{3}{c}{FSB~\cite{chugh2018fingerprint}} & \multicolumn{3}{c}{CFD-PAD~\cite{liu2022fingerprint}} & \multicolumn{3}{c}{PADRD-Net~\cite{fei2024fingerprint}} & \multicolumn{3}{c}{CFM-Net (Ours)} \\
    \hline
    Trainin Sensor & Test Sensor  &ACER  &BER    &TDR   &ACER  &BER   &TDR   &ACER &BER    &TDR  &ACER &BER  &TDR\\
    \hline
    GreenBit & Dermalog     &10.37    & 29.19  &47.43   &12.04    &26.87   &54.87          &6.28       &7.85  &80.72  &\textbf{2.23} &\textbf{4.48} &\textbf{94.95}\\
    Dermalog & GreenBit    &14.82    &17.46   &75.16       &8.50    &10.28   &72.53       &7.95       &8.11  &80.51  &\textbf{5.17} &\textbf{5.47} &\textbf{93.92}\\
    \hline
    \end{tabular}%
\end{center}
  \label{tab:cross sensor}%
\end{table*}%

\begin{table*}[htbp]
  \caption{Performance comparison with previous methods under Cross-Material and Cross-Sensor scenario on LivDet 2021 dataset. BER refers to BPCER@APCER=1\%(\%). TDR refers to TDR@FDR=1\%(\%). The best results are highlighted.}
\setlength{\tabcolsep}{6pt}
\begin{center}
    \begin{tabular}{cccccccccccccc}
    \hline
    \multicolumn{2}{c}{\multirow{2}{*}{LivDet 2021}} & \multicolumn{12}{c}{LivDet2021~~~~ACER(\%)$\downarrow$~~~~BPCER@APCER=1\%(\%)$\downarrow$~~~~TDR@FDR=1\%(\%)$\uparrow$~~~~} \\
\cline{3-14}    \multicolumn{2}{c}{} & \multicolumn{3}{c}{FSB~\cite{chugh2018fingerprint}} & \multicolumn{3}{c}{CFD-PAD~\cite{liu2022fingerprint}} & \multicolumn{3}{c}{PADRD-Net~\cite{fei2024fingerprint}} & \multicolumn{3}{c}{CFM-Net (Ours)} \\
    \hline
    Trainin Sensor & Test Sensor &ACER  &BER    &TDR  &ACER  &BER   &TDR  &ACER &BER    &TDR &ACER &BER    &TDR\\
    \hline
    GreenBit & Dermalog  &26.82    & 90.75  & 43.51  &23.43    &70.52   &46.65   &16.38       &44.04 &65.22 &\textbf{9.87} &\textbf{30.73} &\textbf{68.06}\\
    Dermalog & GreenBit  &21.85    &48.15   &68.57   &36.88    &76.82   &18.82   &11.39       &~\textbf{14.43} &69.08 &\textbf{9.09} & 24.73 & \textbf{71.70}\\
    \hline
    \end{tabular}%
\end{center}
  \label{tab:cross material and sensor}%
\end{table*}%

\textbf{Cross-Sensor Evaluation.} To establish a cross-sensor evaluation protocol, we reorganize the LivDet 2021 dataset by treating each sensor's complete data collection as a subset. Specifically, we employ one sensor's entire dataset for training and validation while reserving another sensor's data exclusively for testing. Based on the results presented in Table~\ref{tab:cross sensor}, our approach outperforms state-of-the-art methods across all metrics. This can be attributed CFM structure also enhances the capacity to capture more comprehensive artifact evidence in regions with varying AED.

\textbf{Cross-Material and Cross-Sensor Evaluation.} To further validate the robustness of our approach, we introduce a more challenging dual-constraint scenario that simultaneously imposes both cross-sensor and cross-material conditions. In this setup, distinct sensors are utilized for training and testing while maintaining identical material distributions as in the cross-material scenario. This evaluation protocol is significantly more demanding, as it requires the model to jointly address generalization challenges arising from both material diversity and sensor variations. As demonstrated in Table~\ref{tab:cross material and sensor}, CFM-Net consistently outperforms the other current methods, underscoring the robustness and effectiveness of our approach under stringent evaluation conditions. A key advantage of our method lies in the interpretability of AED, which provides explicit physical semantics by quantifying local Gabor feature certainty within each region. This interpretable representation enables the model to effectively adapt to previously unseen sensors and materials. The experimental results validate that our approach is well-suited for practical deployment scenarios where simultaneous variations in both acquisition sensors and spoof materials need to be taken into account.

\subsection{Performance on Fingerprints with different AED}
\label{sec: Performance on Fingerprints with Different Qualities}
In this section, we evaluate the performance of our method on fingerprints with different AED values, the result is shown in Table~\ref{tab:abl_qua}. It can be observed that our method achieved the best performance across fingerprints in various ranges of quality. Particularly, for fingerprints with Avg\_AED $\geq$ 0.15, BPCER@APCER=$1\%$ metric improved from 74.64\% to 42.25\%, and TDR@FDR=$1\%$ metric improved from 38.26\% to 78.44\%. This demonstrates that the CFM structure effectively extracts potential artifact evidence from regions with higher AED and integrates the artifact evidence from regions with different AED through multi-depth feature extraction.

\begin{table*}[htbp]
  \caption{The performance on fingerprints with different ranges of average AED of LivDet 2021 dataset.}
\setlength{\tabcolsep}{8pt}
\begin{center}
    \begin{tabular}{cccc}
    \hline
    \multirow{2}{*}{Methods} &\multirow{2}{*}{Test Data} & \multicolumn{2}{c}{LivDet2021} \\
\cline{3-4} 
& &BPCER@APCER=$1\%$(\%)$\downarrow$    &TDR@FDR=$1\%$(\%)$\uparrow$    \\
    \hline
    \multirow{3}{*}{FSB\cite{chugh2018fingerprint}} & Avg\_AED $\leq$ 0.10 &  33.38     & 59.92 \\
     & 0.10 $\textless$ Avg\_AED $\textless$ 0.15 & 43.64 & 58.66\\
     & Avg\_AED $\geq$ 0.15 &74.64 &36.56 \\
     \hline
    \multirow{3}{*}{CFD-PAD~\cite{liu2022fingerprint}} & Avg\_AED $\leq$ 0.10 & 72.87    & 54.79\\
     & 0.10 $\textless$ Avg\_AED $\textless$ 0.15 & 91.90 & 44.25\\
     & Avg\_AED $\geq$ 0.15 & 95.73 & 38.26\\
     \hline
    \multirow{3}{*}{PADRD-Net~\cite{fei2024fingerprint}} & Avg\_AED $\leq$ 0.10 & 23.25      & 83.92 \\
    & 0.10 $\textless$ Avg\_AED $\textless$ 0.15 &28.18 & \textbf{83.36}\\
     & Avg\_AED $\geq$ 0.15 & 77.46 & 27.96\\
     \hline
\multirow{3}{*}{CFM-Net (Ours)} & Avg\_AED $\leq$ 0.10 &\textbf{13.73} & \textbf{89.69} \\
     & 0.10 $\textless$ Avg\_AED $\textless$ 0.15 &\textbf{12.28} & 83.05\\
     & Avg\_AED $\geq$ 0.15 & \textbf{42.25} & \textbf{78.44}\\
    \hline
    \end{tabular}%
\end{center}
    \label{tab:abl_qua}
\end{table*}%

\begin{table*}[htbp]
  \caption{Performance of CFM-Net with or without CFM structure and OGAT module on LivDet 2021 dataset. BER refers to BPCER@APCER=1\%(\%). TDR refers to TDR@FDR=1\%(\%).}
\setlength{\tabcolsep}{8pt}
\begin{center}
    \begin{tabular}{cccccccccccc}
    \hline
      \multirow{3}{*}{\makecell[c]{CFM\\Structure}} &\multirow{3}{*}{\makecell[c]{OGAT\\Module}} &\multicolumn{10}{c}{LivDet2021~~~~BPCER@APCER=1\%(\%)$\downarrow$~~~~TDR@FDR=1\%(\%)$\uparrow$~~~~}\\
\cline{3-12}    
   & &\multicolumn{2}{c}{GreenBit\_CC} &\multicolumn{2}{c}{GreenBit\_SS} &\multicolumn{2}{c}{Dermalog\_CC} &\multicolumn{2}{c}{Dermalog\_SS} &\multicolumn{2}{c}{ALL} \\
\cline{3-12} 
          & &BPCER    &TDR  &BER    &TDR &BER    &TDR &BER    &TDR &BER    &TDR\\
\hline
    \texttimes       &\texttimes       &4.14       &93.53  &26.54 &66.17 &8.29 &93.29 &34.78 &57.17 &34.36 &70.39\\
    \checkmark       &\texttimes       &6.41 &92.64 &18.02 &83.57 &5.80 &93.37 &33.56 &65.13 &19.27 &83.70\\
    \texttimes       &\checkmark       &4.34 &94.78 &\textbf{14.99} &78.27 &6.64 &87.32 &30.97 &57.05 &15.87 &80.05\\
    \checkmark       &\checkmark       &\textbf{3.91} &\textbf{97.30} &16.96 &\textbf{85.99} &\textbf{3.24} &\textbf{97.96} &\textbf{30.73} &\textbf{78.70} &\textbf{14.60} &\textbf{88.91}  \\
    \hline
    \end{tabular}%
\end{center}
  \label{tab:ablation study}%
\end{table*}%

\subsection{Effectiveness of Each Proposed Module}
\label{section:Effectiveness Validation of Each Proposed Module}
In this section, an ablation experiment is conducted to assess the effectiveness of our proposed CFM structure and OGAT module on the LivDet2021 dataset. The results are presented in Table~\ref{tab:ablation study}.

The comparison between the first and second rows demonstrates that our proposed CFM structure accurately captures artifact evidence across regions with varying AED values.
The comparison between the first and third rows reveals that our proposed OGAT module effectively disentangles identity information from PAD features. By leveraging fingerprint orientation, which encodes identity information, as the adversarial training target, the module achieves more thorough identity information removal. Furthermore, conducting adversarial training at the feature map level preserves the integrity of the original artifact evidence.
A comparative analysis of the latter three rows reveals that both the CFM structure and OGAT module independently contribute to performance enhancement, with their combination yielding superior results. This synergistic effect stems from their orthogonal functionalities: the CFM structure enables comprehensive extraction of artifact evidence across regions with varying AED, while the OGAT module filters identity information from PAD features. Their joint operation facilitates the extraction of more robust and discriminative features for the PAD task.

\subsection{The Complementarity of Each Stage's Output Feature}

The experiments in Table~\ref{tab:cfm block} demonstrate the complementarity of each stage's output features within the CFM structure. Specifically, by retaining stages 1, 2, and 3 of the CFM structure and concatenating the output features of each stage for classification, we observe incremental improvements in performance with an increasing number of stages. This indicates that artifact evidence extracted from different regions of fingerprints by the CFM structure exhibits complementary characteristics.

\begin{table}[htbp]
  \caption{The ablation study results to analyze the complementarity of each stage's output feature.}
\setlength{\tabcolsep}{1.5pt}
\begin{center}
    \begin{tabular}{ccc}
    \hline
    \multirow{2}{*}{Utilized Stages} & \multicolumn{2}{c}{LivDet2021} \\
\cline{2-3}      &BPCER@APCER=$1\%$(\%)    &TDR@FDR=$1\%$(\%)    \\
    \hline
    1 & 25.02      & 76.46 \\
    2 &  17.68     & 85.39\\
    3 (Ours) &\textbf{14.60}       &\textbf{88.91} \\
    \hline
    \end{tabular}%
\end{center}
    \label{tab:cfm block}
\end{table}%

\subsection{Analysis of AED design} 

In this section, comparative experiments are conducted to demonstrate the effectiveness of employing local Gabor feature certainty for representing AED. The experimental results are presented in Table~\ref{tab:selection of OCM}. ``AED\_Top1'' represents the direct utilization of the top 1 probability of 90 discrete angles as the AED. ``AED\_Top9'' refers to our proposed AED, which is based on the summation of the top 9 probabilities of 90 discrete angles. ``Adversarial Erasing''~\cite{zhang2018adversarial} is a strategy that forces the network to focus on less-attentional regions by alternately training it on the original images and the regions highlighted in CAM visualization during the training.
The superior performance of ``AED\_Top9'' over ``Adversarial Erasing'' stems from its underlying mechanism. The local Gabor feature certainty derived from fixed-frequency Gabor filters can effectively capture ridge structure clarity while remaining insensitive to internal details. This ensures that regions with similar Gabor values exhibit consistent AED and evidence patterns. Consequently, the CFM applies uniform feature extraction to these homogeneous regions while performing deeper extraction on areas with higher AED values. This hierarchical approach avoids the fundamental limitation of ``Adversarial Erasing" that relies on a rigid feature extractor. Forcing this unified model to shift its focus disrupts the parameters optimized for attentional regions.
The superior performance of ``AED\_Top9'' over ``AED\_Top1'' stems from its enhanced robustness in handling regions with sharp orientation changes, such as those containing minutiae. In these regions, the certainty of the top 1 probability may decrease even though the ridge structure remains clear. This issue is effectively mitigated by adopting the more relaxed operator of summing the top 9 probabilities.

\begin{table}[htbp]
  \caption{The comparison results to analyze the effectiveness of AED design.}
\setlength{\tabcolsep}{1.5pt}
\begin{center}
    \begin{tabular}{ccc}
    \hline
    \multirow{2}{*}{Mehods} & \multicolumn{2}{c}{LivDet2021} \\
\cline{2-3}      &BPCER@APCER=$1\%$(\%)    &TDR@FDR=$1\%$(\%)    \\
    \hline
    Adversarial Erasing~\cite{zhang2018adversarial} & 17.95     &76.12  \\
    AED\_Top1 & 16.46      & 80.26 \\
    AED\_Top9 (Ours) &\textbf{14.60}       &\textbf{88.91} \\
    \hline
    \end{tabular}%
\end{center}
    \label{tab:selection of OCM}
\end{table}%

\subsection{Analysis of CFM structure}

\subsubsection{AED guided Structure Comparison}

In this section, we compare the effectiveness of different architectures in adaptively processing regions with varying AED values. The corresponding results are presented in Table~\ref{tab:AED guided Structure}. ``Constant-depth distinct extractors'' refers to the use of independent feature extractors with equal depth (ResNet-18) for regions with different AED; ``Adaptive-depth distinct extractors'' denotes the employment of independent extractors whose depth is adjusted according to the AED value of each region, and then concatenating the final feature maps from each individual feature extractor and feeding them into the classifier. And ``Cascade extractors'' represents our proposed CFM structure.

The results show that our CFM structure achieves the best performance. This is attributed to its ability to balance efficient cross-region interaction with targeted evidence mining in high-AED regions. Furthermore, owing to the parameter-sharing mechanism in its shallow feature extraction stages, the CFM structure also exhibits the smallest model parameters and the lowest computational cost among the compared architectures.

\begin{table}[htbp]
  \caption{The comparison results to analyze the choice of AED guided Structure.}
\scriptsize
\setlength{\tabcolsep}{0.5pt}
\begin{center}
    \begin{tabular}{ccc}
    \hline
    \multirow{2}{*}{Structure} & \multicolumn{2}{c}{LivDet2021} \\
\cline{2-3}      &BPCER@APCER=$1\%$(\%)    &TDR@FDR=$1\%$(\%)    \\
    \hline
    Constant-depth distinct extractors & 19.77      & 83.39 \\
    Adaptive-depth distinct extractors & 16.33      & 87.88 \\
    Cascade extractors (Ours) &\textbf{14.60}       &\textbf{88.91} \\
    \hline
    \end{tabular}%
\end{center}
    \label{tab:AED guided Structure}
\end{table}%

\subsubsection{Choice of Binarized AEDs Thresholds}

We investigate the threshold selection within the AEDs in the CFM structure. As depicted in Table~\ref{tab:cfm threshold}, we conduct experiments using different thresholds, specifically [0,0,0], [0,0.1,0.15], [0.05,0.1,0.15], and [0,0.05,0.1]. The most optimal performance is achieved when the thresholds are set to [0,0.05,0.1], while comparatively poorer performance is observed with threshold configurations of [0,0.1,0.15] and [0.05,0.1,0.15]. This disparity can be attributed to the overly stringent constraints imposed by the [0,0.1,0.15] threshold configuration, which results in the inability to efficiently extract features within the 0.05$\leq$AED$\leq$0.1 range using only two blocks. The diminished performance observed with threshold values of [0.05,0.1,0.15] is attributed to the exclusion of significant artifact evidence from the 0$\leq$AED$\leq$0.05 regions.

\begin{table}[htbp]
  \caption{The ablation study results to analyze the choice of Binarized AEDs thresholds of CFM structure.}
\setlength{\tabcolsep}{2pt}
\begin{center}
    \begin{tabular}{ccc}
    \hline
    \multirow{2}{*}{Thresholds} & \multicolumn{2}{c}{LivDet2021} \\
\cline{2-3}      &BPCER@APCER=$1\%$(\%)    &TDR@FDR=$1\%$(\%)    \\
    \hline
    0, 0, 0 & 15.87      & 80.05 \\
    0, 0.1, 0.15 &20.90       &76.09 \\
    0.05, 0.1, 0.15 &39.29 &51.59 \\
    0, 0.05, 0.1 (Ours) &\textbf{14.60}       &\textbf{88.91} \\
    \hline
    \end{tabular}%
\end{center}
    \label{tab:cfm threshold}
\end{table}%

\subsubsection{The transferability of CFM structure}

Due to the orthogonality of the CFM structure with existing PAD methods, we aim to demonstrate its universality by evaluating its performance when integrated into other PAD methods. Specifically, the backbone of other PAD methods is divided into three stages, and a binarized AED map consistent with that used in CFM-Net is inserted before each stage. This design ensures that the computation cost does not increase significantly after incorporating the CFM structure, thereby enabling a fairer performance comparison. In this section, we primarily evaluate the impact of integrating the CFM structure into FSB~\cite{chugh2018fingerprint} and CFD-PAD~\cite{liu2022fingerprint}. The results in Table~\ref{tab:orth} indicate that both methods achieve performance improvements upon integrating the CFM structure, demonstrating its stable effectiveness within other frameworks.

\begin{table}[htbp]
  \caption{The ablation study results to analyze the transferability of the CFM structure.}
\setlength{\tabcolsep}{2pt}
\begin{center}
    \begin{tabular}{ccc}
    \hline
    \multirow{2}{*}{Method} & \multicolumn{2}{c}{LivDet2021} \\
\cline{2-3}      &BPCER@APCER=$1\%$(\%)    &TDR@FDR=$1\%$(\%)    \\
    \hline
    FSB\_w/o\_CFM & 36.09      & 68.01 \\
    FSB\_w/ \_CFM & \textbf{30.70}       &\textbf{76.80} \\
    CFD-PAD\_w/o\_CFM & 85.07 & 30.42\\
    CFD-PAD\_w/\_CFM &  \textbf{70.48}     & \textbf{42.67}\\
    \hline
    \end{tabular}%
\end{center}
    \label{tab:orth}
\end{table}%

\subsection{Analysis of OGAT module}
\label{section:Choice of OGAT balance coefficient}

Our proposed OGAT module employs an adversarial reconstruction network guided by orientation to filter out identity information without harming the original artifact evidence. To demonstrate the effectiveness of the OGAT module, we compare it with the MPRR operation used in PADRD-Net~\cite{fei2024fingerprint}. Experimental results are presented in Table~\ref{tab:OGAT and MPRR}. It can be observed that OGAT surpasses MPRR in both BPCER@APCER=1\% and TDR@FDR=1\% metrics, demonstrating its capability to more effectively and comprehensively filter out fingerprint identity information without harming ridge structure, thus preserving the original artifact evidence.

On the other hand, we assess the influence of the balance coefficient $\lambda$ between classification loss and adversarial loss in Table~\ref{tab:coeff}. Based on the experimental results, it can be observed that optimal performance is achieved when $\lambda$ is set to 1. This phenomenon arises because identity information is insufficiently removed when $\lambda$ is too small. Conversely, excessively large values of $\lambda$ lead to the disruption of the PAD information that the main PAD network intends to learn.

\begin{figure*}[tbp]
\centering
   \includegraphics[width=0.85\linewidth]{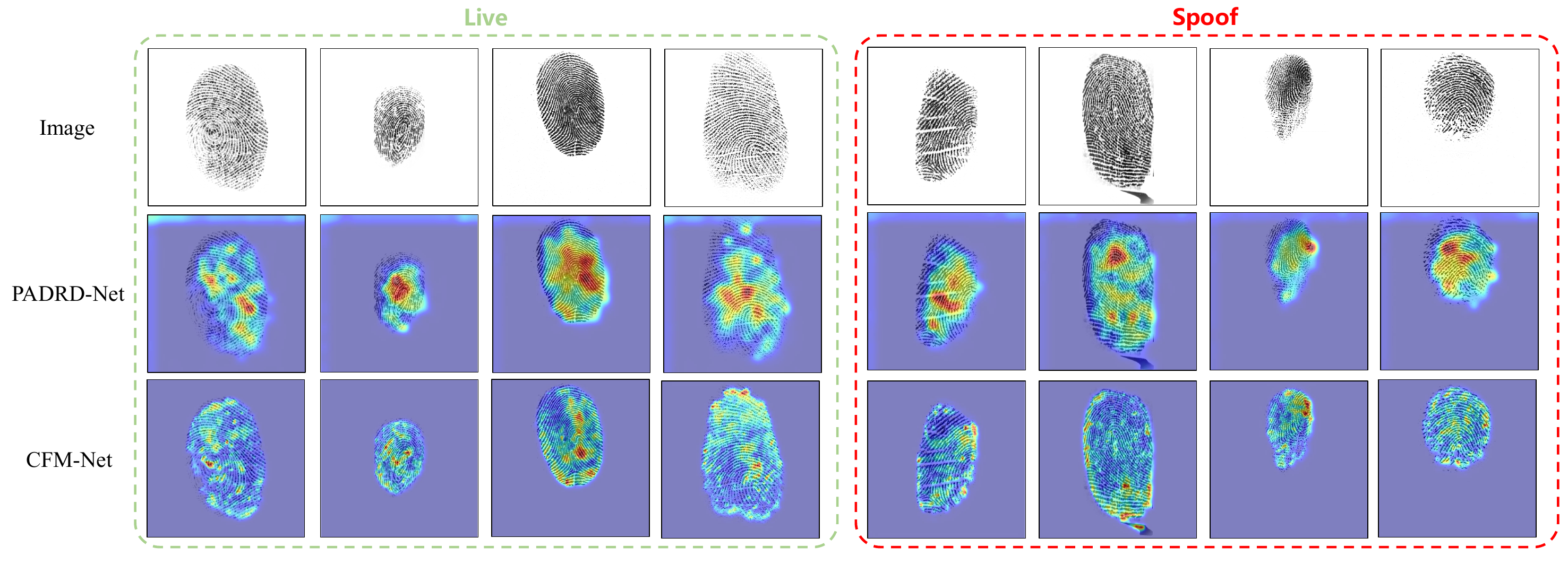}
   \caption{The CAM visualizations on LivDet2021 dataset. The four columns on the left represent live fingerprints, while the four columns on the right represent spoof fingerprints originating from four subsets (GreenBit\_CC, GreenBit\_SS, Dermalog\_CC, Dermalog\_SS). The second row showcases CAM visualizations obtained from PADRD-Net, and the third row showcases CAM visualizations obtained from our proposed CFM-Net.}
\label{fig:cam_vis}
\end{figure*}

\begin{table}[htbp]
\setlength{\tabcolsep}{4pt}
  \caption{The performance comparison between OGAT module and MPRR operation.}
\begin{center}
    \begin{tabular}{ccc}
    \hline
    \multirow{2}{*}{Method} & \multicolumn{2}{c}{LivDet2021} \\
\cline{2-3}      &BPCER@APCER=$1\%$(\%)    &TDR@FDR=$1\%$(\%)    \\
    \hline
    MPRR~\cite{fei2024fingerprint} & 17.68      & 85.75 \\
    OGAT (Ours) &\textbf{14.60}       &\textbf{88.91} \\
    \hline
    \end{tabular}%
\end{center}
    \label{tab:OGAT and MPRR}
\end{table}%

\begin{table}[htbp]
  \caption{The experimental results to analyze the choice of the coefficient of adversarial loss.}
\setlength{\tabcolsep}{4pt}
\begin{center}
    \begin{tabular}{ccc}
    \hline
    \multirow{2}{*}{$\lambda$} & \multicolumn{2}{c}{LivDet2021} \\
\cline{2-3}      &BPCER@APCER=$1\%$(\%)    &TDR@FDR=$1\%$(\%)    \\
    \hline
    0   & 19.27      & 83.70 \\
    0.2 & 21.07      & 80.58 \\
    0.5 & 15.85      & 82.70\\
    1 (Ours) &\textbf{14.60}       &\textbf{88.91} \\
    2 &  19.95               & 74.39       \\
    \hline
    \end{tabular}%
\end{center}
    \label{tab:coeff}
\end{table}%

\subsection{Grad-CAM Visualization}
\label{section:Grad-CAM Visualization}

In this section, we employ the CAM visualization on our proposed model to explain the attention of deep neural networks. The visualization results in Fig.~\ref{fig:cam_vis} depict fingerprints from the LivDet2021 dataset. The left four columns represent live fingerprint samples, and the right four columns represent spoof fingerprint samples from each subset (GreenBit\_CC, GreenBit\_SS, Dermalog\_CC, Dermalog\_SS). The first row shows the original fingerprint images, the second row shows the visualization of PADRD-Net~\cite{fei2024fingerprint}, and the third row shows the visualization of our proposed CFM-Net. It can be observed that, compared to PADRD-Net, CFM-Net can perceive various regions containing artifact evidence rather than solely focusing on partial regions. This further validates the CFM structure's effectiveness in capturing more precise and comprehensive artifact evidence.

\subsection{Limitation Analysis and Future Work}
\label{section:Future Work}
\begin{figure}[htbp]
\centering
   \includegraphics[width=0.65\linewidth]{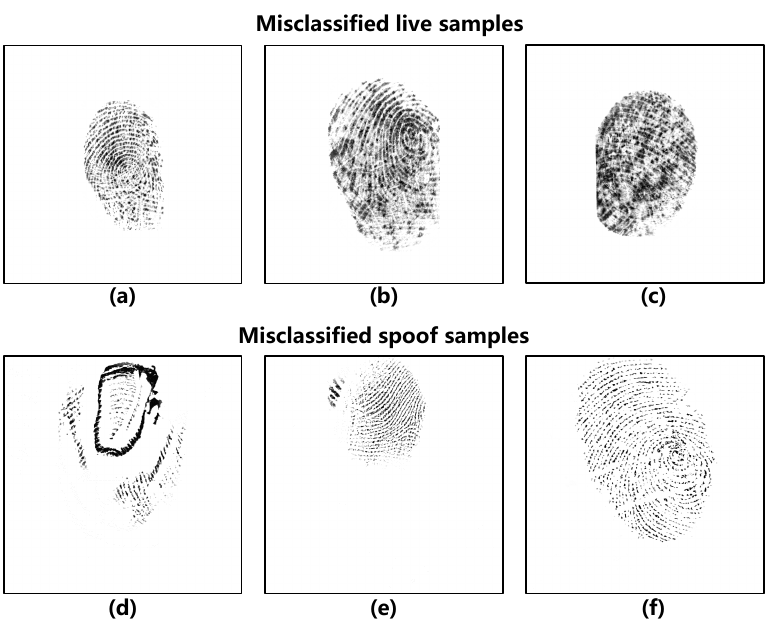}
   \caption{Visualization of misclassified samples. The primary characteristics of these samples are excessively collection environments or insufficient pressure during scanning}
\label{fig:mis_cls}
\end{figure}

Fig.~\ref{fig:mis_cls} shows representative misclassified live/spoof samples. The primary characteristics of these samples are excessively collection environments or insufficient pressure during scanning, which resulted in key artifact evidence, such as sweat secretion, not being captured during this stage. To address these issues, potential improvements in the future include incorporating positive or negative Berlin noise in the data augmentation during the training phase to simulate more complex scanning conditions, thereby enhancing the algorithm's robustness to such samples. Alternatively, during actual use, a re-scanning could be required when the average AED of the input fingerprint is too high, in order to mitigate the impact of unfavorable scanning conditions.

\section{Conclusion}

In this paper, we propose an Artifact Extraction Difficulty (AED) guided Cascade Forgery Mining Network (CFM-Net) for fingerprint presentation attack detection. We first observe that different regions of a fingerprint image can exhibit varying AED, with high-AED regions requiring more sophisticated extraction mechanisms to capture more subtle artifact evidence. To address this issue, we propose using local Gabor feature certainty as a spatial prior to quantify AED, enabling adaptive region-wise processing. Building upon this foundation, CFM-Net employs an adaptive-depth feature extraction architecture to comprehensively mine artifact evidence across regions with varying AED values. Furthermore, an Orientation Guided Adversarial Training (OGAT) module is designed to remove identity information at the feature map level while preserving original artifact evidence. Extensive experiments on the LivDet 2021 and 2019 datasets show that CFM-Net outperforms state-of-the-art methods under cross-material, cross-sensor, and high-AED scenarios. Ablation studies confirm the effectiveness of each proposed component and their complementary roles. In summary, our proposed AED guided CFM structure offers a robust and generalizable solution for fingerprint PAD, enhancing both detection accuracy and generalization. This work provides meaningful insights for future research in fingerprint PAD task.

\bibliographystyle{ieeetr}
\bibliography{main}

@inproceedings{zhou2016learning,
  title={Learning deep features for discriminative localization},
  author={Zhou, Bolei and Khosla, Aditya and Lapedriza, Agata and Oliva, Aude and Torralba, Antonio},
  booktitle={Proceedings of the IEEE conference on computer vision and pattern recognition},
  pages={2921--2929},
  year={2016}
}

@inproceedings{matsumoto2002impact,
  title={Impact of artificial" gummy" fingers on fingerprint systems},
  author={Matsumoto, Tsutomu and Matsumoto, Hiroyuki and Yamada, Koji and Hoshino, Satoshi},
  booktitle={Proceedings of the Optical Security and Counterfeit Deterrence Techniques IV},
  volume={4677},
  pages={275--289},
  year={2002},
  organization={SPIE}
}

@article{arora2016design,
  title={Design and fabrication of 3D fingerprint targets},
  author={Arora, Sunpreet S and Cao, Kai and Jain, Anil K and Paulter, Nicholas G},
  journal={IEEE Transactions on Information Forensics and Security},
  volume={11},
  number={10},
  pages={2284--2297},
  year={2016},
}

@article{arora2017gold,
  title={Gold fingers: 3D targets for evaluating capacitive readers},
  author={Arora, Sunpreet S and Jain, Anil K and Paulter, Nicholas G},
  journal={IEEE Transactions on Information Forensics and Security},
  volume={12},
  number={9},
  pages={2067--2077},
  year={2017},
}

@article{yoon2012altered,
  title={Altered fingerprints: Analysis and detection},
  author={Yoon, Soweon and Feng, Jianjiang and Jain, Anil K},
  journal={IEEE Transactions on Pattern Analysis and Machine Intelligence},
  volume={34},
  number={3},
  pages={451--464},
  year={2012},
}

@article{marasco2014survey,
  title={A survey on antispoofing schemes for fingerprint recognition systems},
  author={Marasco, Emanuela and Ross, Arun},
  journal={ACM Computing Surveys},
  volume={47},
  number={2},
  pages={1--36},
  year={2014},
}

@inproceedings{casula2021livdet,
  title={Livdet 2021 fingerprint liveness detection competition-into the unknown},
  author={Casula, Roberto and Micheletto, Marco and Orr{\`u}, Giulia and Delussu, Rita and Concas, Sara and Panzino, Andrea and Marcialis, Gian Luca},
  booktitle={Proceedings of the International Joint Conference on Biometrics},
  pages={1--6},
  year={2021},
  publisher={IEEE}
}

@inproceedings{lee2009fake,
  title={Fake finger detection using the fractional Fourier transform},
  author={Lee, Hyun-suk and Maeng, Hyun-ju and Bae, You-suk},
  booktitle={Proceedings of Biometric ID Management and Multimodal Communication: Joint COST 2101 and 2102 International Conference},
  pages={318--324},
  year={2009},
  publisher={Springer}
}

@article{zhang2019slim,
  title={Slim-ResCNN: A deep residual convolutional neural network for fingerprint liveness detection},
  author={Zhang, Yongliang and Shi, Daqiong and Zhan, Xiaosi and Cao, Di and Zhu, Keyi and Li, Zhiwei},
  journal={IEEE Access},
  volume={7},
  pages={91476--91487},
  year={2019},
}

@inproceedings{nikam2008fingerprint,
  title={Fingerprint liveness detection using curvelet energy and co-occurrence signatures},
  author={Nikam, Shankar Bhausaheb and Agarwal, Suneeta},
  booktitle={Proceedings of the International Conference on Computer Graphics, Imaging and Visualisation},
  pages={217--222},
  year={2008},
  publisher={IEEE}
}

@article{xia2018novel,
  title={A novel weber local binary descriptor for fingerprint liveness detection},
  author={Xia, Zhihua and Yuan, Chengsheng and Lv, Rui and Sun, Xingming and Xiong, Neal N and Shi, Yun-Qing},
  journal={IEEE Transactions on Systems, Man, and Cybernetics: Systems},
  volume={50},
  number={4},
  pages={1526--1536},
  year={2018},
}

@article{antonelli2006fake,
  title={Fake finger detection by skin distortion analysis},
  author={Antonelli, Athos and Cappelli, Raffaele and Maio, Dario and Maltoni, Davide},
  journal={IEEE Transactions on Information Forensics and Security},
  volume={1},
  number={3},
  pages={360--373},
  year={2006},
}

@article{liu2022fingerprint,
  title={Fingerprint presentation attack detection by channel-Wise feature denoising},
  author={Liu, Feng and Kong, Zhe and Liu, Haozhe and Zhang, Wentian and Shen, Linlin},
  journal={IEEE Transactions on Information Forensics and Security},
  volume={17},
  pages={2963--2976},
  year={2022},
}

@article{lapsley1998anti,
  title={Anti-fraud biometric scanner that accurately detects blood flow},
  author={Lapsley, Philip Dean and Lee, Jonathan Alexander and Pare Jr, David Ferrin and Hoffman, Ned},
  journal={Google Patents},
  year={1998},
  note={US Patent 5,737,439}
}

@inproceedings{baldisserra2005fake,
  title={Fake fingerprint detection by odor analysis},
  author={Baldisserra, Denis and Franco, Annalisa and Maio, Dario and Maltoni, Davide},
  booktitle={Proceedings of the International Conference on Biometrics},
  pages={265--272},
  year={2005},
  publisher={IEEE}
}

@article{engelsma2018raspireader,
  title={Raspireader: Open source fingerprint reader},
  author={Engelsma, Joshua J and Cao, Kai and Jain, Anil K},
  journal={IEEE Transactions on Pattern Analysis and Machine Intelligence},
  volume={41},
  number={10},
  pages={2511--2524},
  year={2018},
}

@article{cheng2006artificial,
  title={Artificial fingerprint recognition by using optical coherence tomography with autocorrelation analysis},
  author={Cheng, Yezeng and Larin, Kirill V},
  journal={OPG Applied Optics},
  volume={45},
  number={36},
  pages={9238--9245},
  year={2006},
}

@article{chugh2018fingerprint,
  title={Fingerprint spoof buster: Use of minutiae-centered patches},
  author={Chugh, Tarang and Cao, Kai and Jain, Anil K},
  journal={IEEE Transactions on Information Forensics and Security},
  volume={13},
  number={9},
  pages={2190--2202},
  year={2018},
}

@inproceedings{lekshmy2022one,
  title={One-Shot sensor and material translator: A bilinear decomposer for fingerprint presentation attack generalization},
  author={Lekshmy, Gowri and Namboodiri, Anoop},
  booktitle={Proceedings of the International Joint Conference on Biometrics},
  pages={1--10},
  year={2022},
  publisher={IEEE}
}

@inproceedings{orru2019livdet,
  title={Livdet in action-fingerprint liveness detection competition 2019},
  author={Orr{\`u}, Giulia and Casula, Roberto and Tuveri, Pierluigi and Bazzoni, Carlotta and Dessalvi, Giovanna and Micheletto, Marco and Ghiani, Luca and Marcialis, Gian Luca},
  booktitle={Proceedings of the International Joint Conference on Biometrics},
  pages={1--6},
  year={2019},
  publisher = {IEEE},  
}

@article{sun2023new,
  author={Sun, Haohao and Zhang, Yilong and Chen, Peng and Wang, Haixia and Liu, Yi-Peng and Liang, Ronghua},
  journal={IEEE Transactions on Information Forensics and Security}, 
  title={A New Approach in Automated Fingerprint Presentation Attack Detection Using Optical Coherence Tomography}, 
  year={2023},
  volume={18},
  number={},
  pages={4243-4257},
  }

@article{zhang2024uniform,
  title={A uniform representation model for OCT-based fingerprint presentation attack detection and reconstruction},
  author={Zhang, Wentian and Liu, Haozhe and Liu, Feng and Ramachandra, Raghavendra},
  journal={Elsevier Pattern Recognition},
  volume={145},
  pages={109981},
  year={2024},
  publisher={}
}

@inproceedings{Gajawada2019UniversalMT,
  title={Universal Material Translator: Towards Spoof Fingerprint Generalization},
  author={Rohit Gajawada and Additya Popli and T. Chugh and Anoop M. Namboodiri and Anil K. Jain},
  booktitle={Proceedings of the International Conference on Biometrics},
  pages={1-8},
  year={2019},
  publisher = {IEEE},
}

@article{liu2021fingerprint,
  title={Fingerprint presentation attack detector using global-local model},
  author={Liu, Haozhe and Zhang, Wentian and Liu, Feng and Wu, Haoqian and Shen, Linlin},
  journal={IEEE Transactions on Cybernetics},
  volume={52},
  number={11},
  pages={12315--12328},
  year={2021},
  publisher={IEEE}
}

@inproceedings{huang2023fingerprint,
  title={Fingerprint Presentation Attack Detection with Supervised Contrastive Learning},
  author={Huang, Chuanwei and Fei, Hongyan and Wu, Song and Wang, Zheng and Jia, Zexi and Feng, Jufu},
  booktitle={International Joint Conference on Biometrics},
  pages={1--10},
  year={2023},
  organization={IEEE}
}

@article{fei2024fingerprint,
  title={Fingerprint Presentation Attack Detection by Region Decomposition},
  author={Fei, Hongyan and Huang, Chuanwei and Wu, Song and Wang, Zheng and Jia, Zexi and Feng, Jufu},
  journal={IEEE Transactions on Information Forensics and Security},
  volume={19},
  number={},
  pages={3974-3985},
  year={2024},
  publisher={IEEE}
}

@inproceedings{ganin2015unsupervised,
  title={Unsupervised domain adaptation by backpropagation},
  author={Ganin, Yaroslav and Lempitsky, Victor},
  booktitle={International Conference on Machine Learning},
  pages={1180--1189},
  year={2015},
  organization={PMLR}
}

@inproceedings{goodfellow2014generative,
  title={Generative adversarial nets},
  author={Goodfellow, Ian and Pouget-Abadie, Jean and Mirza, Mehdi and Xu, Bing and Warde-Farley, David and Ozair, Sherjil and Courville, Aaron and Bengio, Yoshua},
  booktitle={Advances in Neural Information Processing Systems},
  pages={2672–2680},
  year={2014},
  organization={MIT Press}
}

@article{ganin2016domain,
  title={Domain-adversarial training of neural networks},
  author={Ganin, Yaroslav and Ustinova, Evgeniya and Ajakan, Hana and Germain, Pascal and Larochelle, Hugo and Laviolette, Fran{\c{c}}ois and March, Mario and Lempitsky, Victor},
  journal={MIT Press Journal of Machine Learning Research},
  volume={17},
  number={59},
  pages={1--35},
  year={2016}
}

@inproceedings{tzeng2017adversarial,
  title={Adversarial discriminative domain adaptation},
  author={Tzeng, Eric and Hoffman, Judy and Saenko, Kate and Darrell, Trevor},
  booktitle={Conference on Computer Vision and Pattern Recognition},
  pages={7167--7176},
  year={2017},
  organization={IEEE}
}

@inproceedings{zhang2018mitigating,
  title={Mitigating unwanted biases with adversarial learning},
  author={Zhang, Brian Hu and Lemoine, Blake and Mitchell, Margaret},
  booktitle={Association for the Advancement of Artificial Intelligence},
  pages={335--340},
  year={2018},
  organization={ACM}
}

@inproceedings{grosz2020fingerprint,
  title={Fingerprint presentation attack detection: A sensor and material agnostic approach},
  author={Grosz, Steven A and Chugh, Tarang and Jain, Anil K},
  booktitle={International Joint conference on Biometrics},
  pages={1--10},
  year={2020},
  organization={IEEE}
}

@inproceedings{zhang2018adversarial,
  title={Adversarial complementary learning for weakly supervised object localization},
  author={Zhang, Xiaolin and Wei, Yunchao and Feng, Jiashi and Yang, Yi and Huang, Thomas S},
  booktitle={Proceedings of the IEEE conference on computer vision and pattern recognition},
  pages={1325--1334},
  year={2018}
}

@article{hamamoto1998gabor,
  title={A Gabor filter-based method for recognizing handwritten numerals},
  author={Hamamoto, Yoshihiko and Uchimura, Shunji and Watanabe, Masanori and Yasuda, Tetsuya and Mitani, Yoshihiro and Tomita, Shingo},
  journal={Pattern recognition},
  volume={31},
  number={4},
  pages={395--400},
  year={1998},
  publisher={Elsevier}
}

@article{gabor1946theory,
  title={Theory of communication. Part 1: The analysis of information},
  author={Gabor, Dennis},
  journal={Journal of the Institution of Electrical Engineers-part III: radio and communication engineering},
  volume={93},
  number={26},
  pages={429--441},
  year={1946},
  publisher={IET}
}

@article{jain1997line,
  title={On-line fingerprint verification},
  author={Jain, Anil and Hong, Lin and Bolle, Ruud},
  journal={IEEE transactions on pattern analysis and machine intelligence},
  volume={19},
  number={4},
  pages={302--314},
  year={1997},
  publisher={IEEE}
}

@article{hong1998fingerprint,
  title={Fingerprint image enhancement: algorithm and performance evaluation},
  author={Hong, Lin and Wan, Yifei and Jain, Anil},
  journal={IEEE transactions on pattern analysis and machine intelligence},
  volume={20},
  number={8},
  pages={777--789},
  year={1998},
  publisher={IEEE}
}

@article{yook2024attention,
  title={Attention Map Is All We Need for Lightweight Fingerprint Liveness Detection},
  author={Yook, Hyun Jun and Hong, Pyo Min and Kang, So Hyun and San Jhun, Ga and Seo, Jae Eun and Lee, Youn Kyu},
  journal={IEEE Access},
  year={2024},
  publisher={IEEE}
}

@article{reza2025cross,
  title={Cross-sensor Generalization for Fingerprint Presentation Attack Detection Leveraging Local Feature Enhancement},
  author={Reza, Naim and Jung, Ho Yub},
  journal={IEEE Transactions on Biometrics, Behavior, and Identity Science},
  year={2025},
  publisher={IEEE}
}

@article{rai2024open,
  title={An open patch generator based fingerprint presentation attack detection using generative adversarial network},
  author={Rai, Anuj and Anshul, Ashutosh and Jha, Ashwini and Jain, Prayag and Sharma, Ram Prakash and Dey, Somnath},
  journal={Multimedia Tools and Applications},
  volume={83},
  number={9},
  pages={27723--27746},
  year={2024},
  publisher={Springer}
}

\vspace{11pt}

\vfill

\end{document}